\documentclass[runningheads]{llncs}
\usepackage[T1]{fontenc}
\usepackage{amsmath}
\usepackage{multirow}
\usepackage{booktabs}
\usepackage{tabularx}
\usepackage{graphicx}
\begin{document}
\title{Progressive Deep Learning for Automated Spheno-Occipital Synchondrosis Maturation Assessment}
%
%
%


\author{
Omid Halimi Milani\inst{1} \and
Amanda Nikho\inst{2} \and
Marouane Tliba\inst{1,3} \and
Lauren Mills\inst{2} \and
Emadeldeen Hamdan\inst{1} \and
Ahmet Enis Cetin\inst{1} \and
Mohammed H. Elnagar\inst{2}
}

\authorrunning{O. H. Milani et al.}

\institute{
Department of Electrical and Computer Engineering, 
University of Illinois Chicago, Chicago, IL, USA
\and
Department of Orthodontics, College of Dentistry, 
University of Illinois Chicago, Chicago, IL, USA
\and
University of Orléans, Orléans, France
}

\maketitle              
\begin{abstract}
Accurate assessment of spheno-occipital synchondrosis (SOS) maturation is a key indicator of craniofacial growth and a critical determinant for orthodontic and surgical timing. However, SOS staging from cone-beam CT (CBCT) relies on subtle, continuously evolving morphological cues, leading to high inter-observer variability and poor reproducibility, especially at transitional fusion stages. We frame SOS assessment as a fine-grained visual recognition problem and propose a progressive representation-learning framework that explicitly mirrors how expert clinicians reason about synchondral fusion: from coarse anatomical structure to increasingly subtle patterns of closure. Rather than training a full-capacity network end-to-end, we sequentially grow the model by activating deeper blocks over time, allowing early layers to first encode stable cranial base morphology before higher-level layers specialize in discriminating adjacent maturation stages. This yields a curriculum over network depth that aligns deep feature learning with the biological continuum of SOS fusion. Extensive experiments across convolutional and transformer-based architectures show that this expert-inspired training strategy produces more stable optimization and consistently higher accuracy than standard training, particularly for ambiguous intermediate stages. Importantly, these gains are achieved without changing network architectures or loss functions, demonstrating that training dynamics alone can substantially improve anatomical representation learning.
The proposed framework establishes a principled link between expert dental intuition and deep visual representations, enabling robust, data-efficient SOS staging from CBCT and offering a general strategy for modeling other continuous biological processes in medical imaging.

\end{abstract}

%
%
%

\section{Introduction}

Understanding biological maturation from medical images is a fundamental challenge in computer vision for healthcare, requiring the extraction of subtle, continuously evolving anatomical patterns from high-dimensional visual data. In orthodontics and dentofacial orthopedics, accurate assessment of craniofacial skeletal maturation directly determines the timing of orthopedic appliances, orthognathic surgery, and growth-modifying interventions. Chronological age is a poor proxy for biological development, with large inter-individual discrepancies driven by genetics, hormonal variation, and growth dynamics \cite{Mito2002,calfee2010skeletal,karlberg2002secular,de2014hand}. As a result, clinicians rely on imaging-based maturity indicators that encode biological progression through subtle morphological changes.
The increasing clinical adoption of cone-beam computed tomography (CBCT) for impacted teeth, asymmetry analysis, and orthognathic planning has made high-resolution three-dimensional visualization of craniofacial growth centers widely available \cite{Scarfe2017,OlchAlaei2021}. Among these structures, the spheno-occipital synchondrosis (SOS) plays a central biological role as a primary growth site of the cranial base, with its progressive fusion tightly linked to pubertal development and maxillofacial growth \cite{alhazmi2021correlation,booth2024correlations,fernandez2016spheno,powell1963closure,lottering2015ontogeny,krishan2013evaluation,tashayyodi2023relationship,alhazmi2017timing,leonardi2010rapid,leonardi2021three,goldstein2014earlier,tahiri2014spheno}. Importantly, SOS maturation can be assessed even when other skeletal indicators fall outside the CBCT field of view, making it a uniquely valuable biomarker for growth estimation \cite{OlchAlaei2021,Kapila2015,alhazmi2021correlation,fernandez2016spheno}.

From a visual recognition perspective, SOS staging is a fine-grained and ordinal classification problem over anatomical structures whose appearance changes continuously rather than discretely. Expert radiologists do not identify SOS stages by detecting isolated landmarks, but by integrating weak morphological cues, such as : endocranial fusion, trabecular continuity, and residual scar patterns—across multiple spatial scales. This makes manual staging both labor-intensive and inherently variable, particularly at transitional fusion stages where inter-class boundaries are visually ambiguous \cite{booth2024correlations,tashayyodi2023relationship}. These characteristics position SOS analysis as a challenging testbed for representation learning in medical imaging, where success depends on capturing both global cranial base geometry and subtle stage-specific textural patterns.

Recent advances in deep learning have enabled strong performance on a wide range of medical image classification tasks, including skeletal and craniofacial analysis \cite{lee2017deep,MohammadRahimi2021,zhang2020automatic,thurzo2021use}. However, prior work on SOS staging from CBCT highlights persistent limitations: performance is highly sensitive to dataset size, class imbalance, and the visual similarity of adjacent maturation stages \cite{booth2024correlations,kumar2024cbct}. For example, fully automated CVM staging from cephalometric radiographs has been explored using CNN-based pipelines with directional edge-enhancing filters \cite{AticiCVMDirectionalFilters}.
 High-capacity convolutional and transformer-based models trained in a single end-to-end regime often struggle to organize SOS morphology into stable, discriminative feature hierarchies under these conditions, leading to unstable convergence and poor generalization; especially for clinically critical transitional stages. A key strength of this study is the use of a rigorously curated, expert-annotated CBCT dataset. All SOS stages were labeled by experienced orthodontists and oral and maxillofacial radiologists following a standardized five-stage fusion protocol, with high inter-rater reliability. Unlike large-scale datasets with weak or noisy supervision, this collection emphasizes anatomical fidelity and diagnostic consistency, ensuring that the learned representations reflect true biological transitions rather than spurious correlations. Such expert-driven annotation is essential for modeling SOS maturation, where subtle visual cues encode clinically meaningful developmental information.

In this work, we propose a progressive training paradigm that bridges expert dental reasoning with deep representation learning for SOS maturation assessment. Inspired by how clinicians first recognize gross cranial base structure before focusing on subtle fusion patterns, we train networks in a coarse-to-fine manner by gradually activating deeper layers during learning. This progressive organization encourages early layers to capture stable anatomical cues, while later layers specialize in discriminating adjacent maturation stages, aligning optimization with the biological continuum of synchondral fusion.

Across CNN and ViT backbones, this simple schedule stabilizes optimization and improves accuracy over standard end-to-end training. This clean comparison is important: it isolates training dynamics as a key factor in learning clinically meaningful SOS representations.

In this work, we introduce a progressive-growing training strategy to improve both accuracy and computational efficiency for SOS maturation assessment. The proposed approach trains the network by gradually activating deeper blocks, stabilizing optimization and enabling more structured feature learning without modifying architectures or loss functions. Experiments across CNN and ViT backbones show consistent gains over standard end-to-end training while reducing overall computation, and we further validate its generality with improved ViT-B/16 performance on CIFAR-10.

\subsection*{Dataset and Expert Annotation}

We conduct a retrospective study using de-identified cone-beam CT (CBCT) volumes collected from multiple private dental clinics across the Midwestern United States under IRB-exempt approval (University of Illinois Chicago OPRS, Study ID: STUDY2022-1048). All scans were acquired on i-CAT CBCT systems using standardized protocols with 120 kV tube voltage, exposure times between 4 and 8 seconds, voxel sizes between 0.25 and 0.40 mm, and extended field-of-view coverage.
Subjects were positioned in natural head posture with stabilized occlusion and relaxed facial musculature. Exclusion criteria removed cases with craniofacial syndromes, head or neck trauma, or prior surgical intervention. The initial archive contained 1200 CBCT volumes spanning ages 7 to 76 years. To ensure high-fidelity supervision for representation learning, scans with ambiguous or liminal SOS fusion patterns, such as weak endocranial demarcation or indistinct scar remnants, were excluded. This resulted in a curated dataset of 723 volumes with ages 7 to 68 years, including 370 female, 260 male, and 93 unspecified-sex subjects.
SOS maturation was labeled by two orthodontists and one oral and maxillofacial radiologist following the five-stage fusion protocol of Bassed et al.~\cite{bassed2010analysis}, which models synchondral closure progressing from the endocranial to the ectocranial margins. Inter-rater reliability was evaluated on 97 scans, yielding a Cronbach’s $\alpha=0.945$, indicating excellent agreement.

Table~\ref{tab:sos_dataset} summarizes the distributions of SOS stages and age cohorts. The dataset exhibits class imbalance and fine-grained inter-stage transitions, which together make SOS staging a challenging setting for learning discriminative anatomical representations.
\begin{table}[t]
\centering
\caption{Expert-curated SOS CBCT dataset. We report 723 annotated scans across SOS maturation stages and subject age cohorts, highlighting class imbalance and broad age coverage.}

\label{tab:sos_dataset}
\setlength{\tabcolsep}{6pt}
\begin{tabular}{lcc|lcc}
\hline
\multicolumn{3}{c|}{\textbf{SOS Stage}} & \multicolumn{3}{c}{\textbf{Age Group}} \\
\textbf{Stage} & \textbf{\#} & \textbf{\%} & \textbf{Group} & \textbf{\#} & \textbf{\%} \\
\hline
Stage 1 & 159 & 22.0 & Child (5 to 12) & 281 & 38.9 \\
Stage 2 & 92  & 12.7 & Adolescent (13 to 19) & 168 & 23.2 \\
Stage 3 & 92  & 12.7 & Adult (20 to 39) & 116 & 16.0 \\
Stage 4 & 125 & 17.3 & Middle Age (40 to 59) & 55 & 7.6 \\
Stage 5 & 255 & 35.3 & Senior Adult (60+) & 9 & 1.2 \\
 &  &  & Unknown & 94 & 13.0 \\
\hline

\end{tabular}
\end{table}

\begin{figure}[t]
\centering
\begin{minipage}[t]{0.49\textwidth}
    \centering
    \includegraphics[height=4.2cm,width=\linewidth,keepaspectratio]{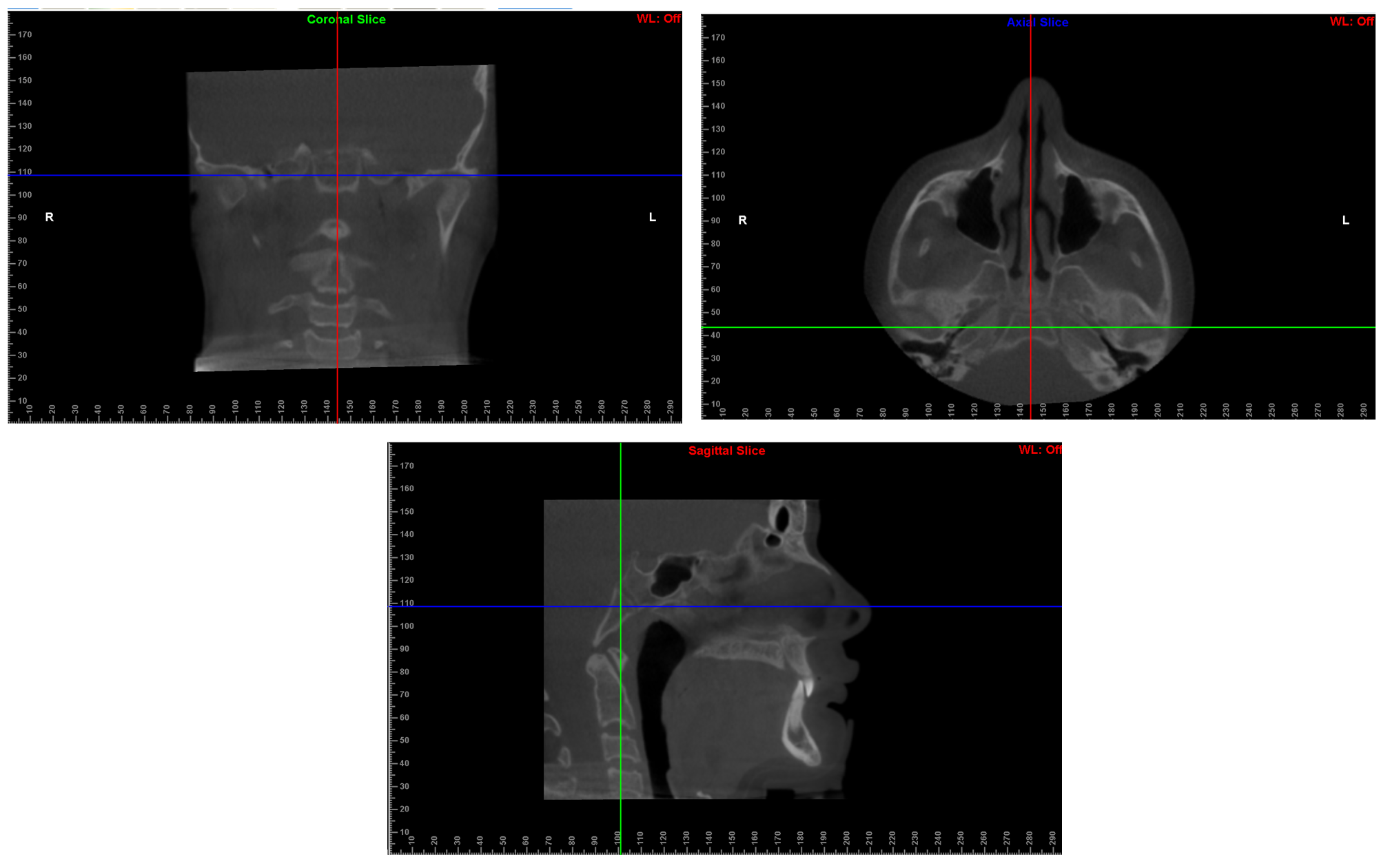}
    \caption{Standard skull alignment across three orthogonal planes used for SOS assessment.}
    \label{fig:orientation}
\end{minipage}
\hfill
\begin{minipage}[t]{0.49\textwidth}
    \centering
    \includegraphics[height=4.2cm,width=\linewidth,keepaspectratio]{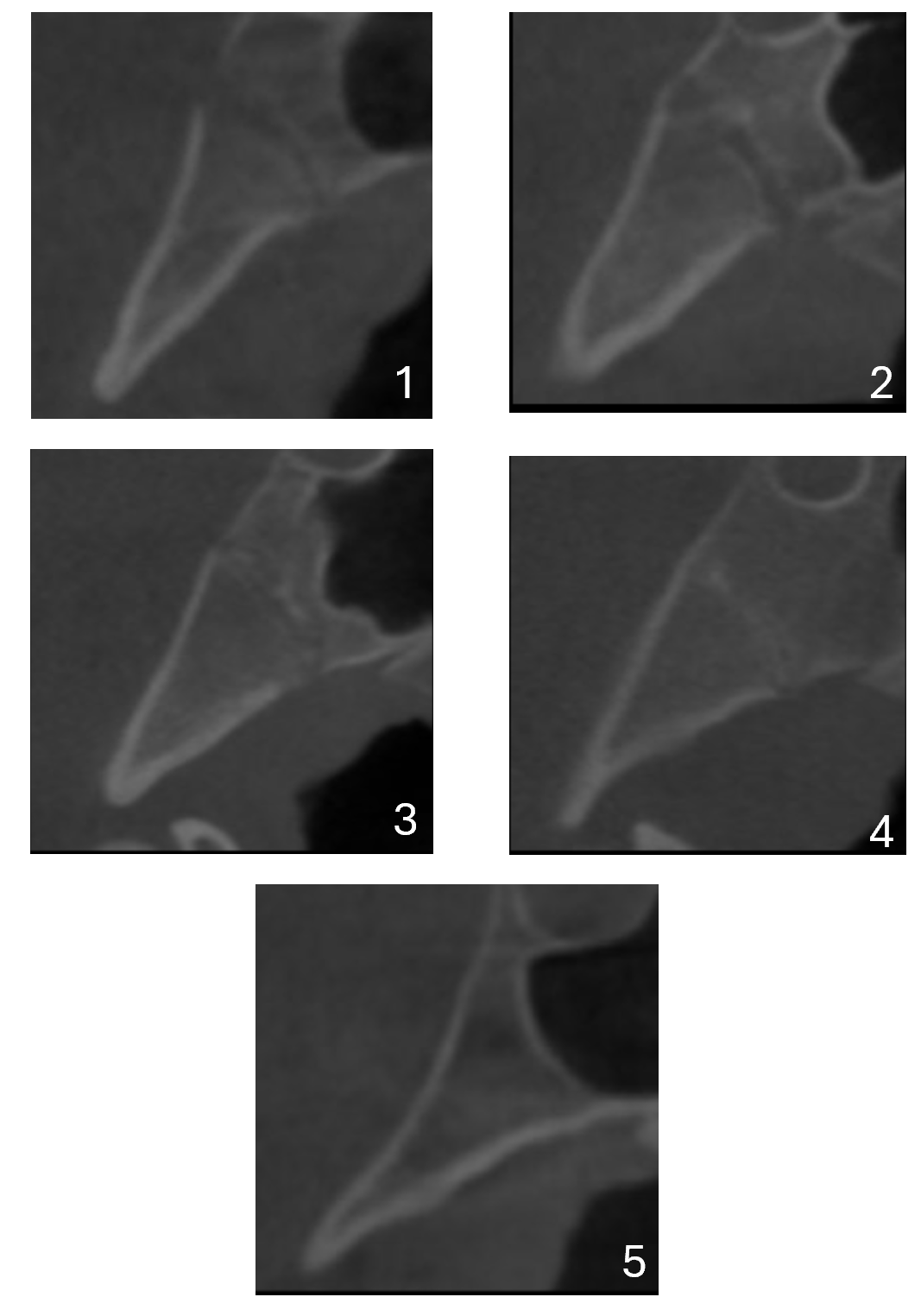}
    \caption{Five-stage SOS maturation scheme used for clinical staging (Stages 1--5).}
    \label{fig:sos_stages}
\end{minipage}
\end{figure}

All CBCT volumes were imported in DICOM format and resampled to a uniform isotropic voxel size of 0.5 mm. Each scan was rigidly reoriented into a standardized cranial reference frame by aligning the basilar occipital plane in coronal and axial views, followed by midline sagittal alignment. From this canonical pose, a single sagittal slice capturing the full spheno-occipital synchondrosis was extracted for analysis. This protocol reduces nuisance variability and ensures that visual differences predominantly reflect anatomical maturity rather than pose or scale artifacts. The alignment procedure and SOS staging scheme are illustrated in Figures~\ref{fig:sos_stages}, and~\ref{fig:orientation}.


\section{Progressive-Growing Training Strategy}

Most adopted Convolutional and Transformer-based backbones such as ResNet and ViT contain tens of millions of parameters, making direct end-to-end optimization brittle and data-inefficient when training sets are limited, as is typical in medical imaging. Rather than exposing the full model capacity from the outset, we adopt a progressive-growing training strategy that explicitly structures how representations are formed over time.

\begin{figure}[t]
    \centering
    \includegraphics[width=0.8\linewidth]{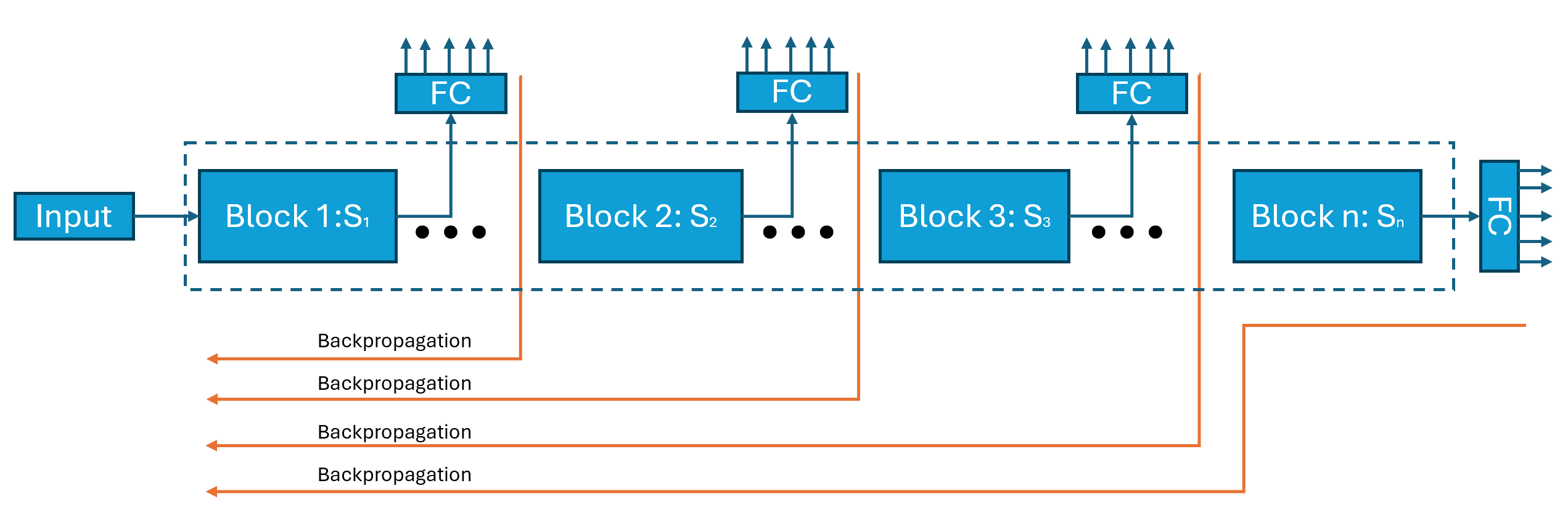}
    \caption{Progressive-growing training. The backbone is expressed as an ordered sequence of blocks. At each stage, a single classification head is attached to the currently active depth and trained. Earlier heads are removed as deeper blocks are activated and training continues. \textit{At inference time, the final fully grown network is used with a single forward pass.}}

    \label{fig:progressive_blocks}
\end{figure}

The network backbone is decomposed into an ordered sequence of blocks (Figure~\ref{fig:progressive_blocks}), and training proceeds by gradually activating deeper blocks. At early stages, only a shallow prefix of the network is trained, with predictions produced by a lightweight classification head. As training progresses, additional blocks are introduced and jointly optimized, reusing the weights learned at previous stages. This continues until the full backbone is active, yielding a standard network for inference.

\paragraph{\textbf{Optimization efficiency.}}
Early stages operate on shallow networks with small activation volumes, substantially reducing memory usage and computation per iteration. This allows the model to learn stable low-level anatomical representations before higher-capacity layers are introduced. As a result, progressive growing achieves comparable or higher final accuracy than full-depth training while requiring significantly fewer parameter updates overall. In practice, training a half-depth network for 50 epochs followed by a full network for 350 epochs requires less than 60\% of the compute of a single 400-epoch end-to-end run, yet converges more reliably.

\paragraph{\textbf{Representation stability.}}
Progressive growing also improves optimization stability. Each newly added block is initialized from a network that already encodes meaningful SOS anatomy, which reduces gradient noise and prevents early overfitting in deep layers. This induces an implicit curriculum over network depth, where low-level features first capture coarse cranial base structure and higher-level features subsequently specialize in discriminating fine-grained fusion stages. The resulting representations are more structured and better aligned with the biological progression of synchondral maturation.

\subsection{Progressive ResNet Stage Partitioning}

Let us consider the standard ResNet architecture $y=R(x)$, where $x$ is an input CBCT image and the output $y \in \{1,2,...,5\}$ covering the five SOS classes.

The backbone consists of a stem
followed by a sequence of $N$ residual blocks:
\begin{equation}
\mathcal{B} = (b_1, b_2, \ldots, b_N),
\end{equation}
where $N = 8$ for ResNet--18, $16$ for ResNet--34 and ResNet--50,
$33$ for ResNet--101, and $50$ for ResNet--152.
Each block $b_i$ is either a BasicBlock (18/34) or a Bottleneck block (50/101/152).
The entire network can be described as follows $R(x)= b_N (b_{N-1} (... b_1(x)...) $.

To enable progressive network training, we partition the ordered blocks of ResNet
 $\mathcal{B}$ into $K \ge N$ stages :
\[
\mathcal{S}_1, \mathcal{S}_2, \ldots, \mathcal{S}_K,
\]
where each stage contains a set of residual blocks.
\[
\mathcal{S}_k = \{ b_{i}, i \in I_k \},
\]
where the index sets $I_k$ form an exact, non-overlapping partition of
$\{1,\dots,N\}$ and preserve the original ordering:
\[
I_1 = \{1,\dots,n_1\},\quad
I_2 = \{n_1{+}1,\dots,n_2\}, \ldots,
I_K = \{n_{K-1}{+}1,\dots,N\}.
\]
Therefore, we have
\begin{equation}
    R(x)= S_K (S_{K-1} (... S_1(x)...)
\end{equation}
where the final block $S_K$ has a fully connected layer with six outputs corresponding to the five SOS classes.

In progressive training,
we create a sequence of networks, as shown in Fig. 1, during training. We start training the network with the first partition
\begin{equation}
y_1= FC_1 ( S_1 (x))
\end{equation}
where $FC_1$ is a fully connected layer with a softmax nonlinearity whose input is the output of $S_1 (x)$ and whose output is $y_1 \in { 1, 2, ...,5}$. We need the fully connected layer to train the parameters of $S_1$ from the available SOS training data. Initial parameters are random numbers.

After we finish training this network, we discard $FC_1$, but we keep the trained parameters of $S_1$ and connect it to  the second stage to obtain a deeper network, i.e.,
\begin{equation}
y_2= FC_2 ( S_2(S_1 (x)))
\end{equation}
where $FC_2$ is another fully connected layer with a softmax nonlinearity, and the output $y_2 \in { 1, 2, ...,5}$ so that we can train $S_2$ using the available SOS training data. Parameters of $S_2$ are initialized with random numbers. We connect the trained $S_2$ to $S_3$ and discard $FC_2$. We continue this process until the last stage. In the last stage, we train the full network  $y=R(x)= S_K (S_{K-1} (... S_1(x)...)$ similar to the regular backpropagation-type training.

It appears that this progressive training approach requires more computation because we train the entire ResNet in the last stage. However, the number of epochs required to train the entire ResNet is significantly lower than training the ResNet from scratch, as discussed in Section 3. Our training approach is essentially a smart parameter initialization method since the training process extracts the morphological features of SOS images in block $S_1$ and they are refined in subsequent stages.

We also call this approach the curriculum training because we first learn Chapter 1 (block $S_1)$, then Chapter 2 (block $S_2)$, ..., and finally the last Chapter K (block $ S_K)$ and the entire course in a sequential manner. In the last step, we revise all the parameters that we have learned by training the entire ResNet.


\vspace{0.4em}
\noindent\textbf{Scenario 1: Four-stage progressive curriculum ($K=4$).}
For a requested four-stage decomposition, the block sequence is divided into four
contiguous groups of \emph{exact} sizes $(n_1,n_2,n_3,n_4)$, computed as the lexicographically
smallest integer partition whose elements differ by at most one:
\[
(n_1,n_2,n_3,n_4) =
\text{BalancedPartition}(N,4).
\]

The exact partitions for all ResNet variants are shown in
Table~\ref{tab:partition-four}.

\begin{table}
\caption{Exact four-stage partitioning used in Scenario~1.}
\label{tab:partition-four}
\begin{tabular}{|c|c|c|}
\hline
Model & $N$ (blocks) & Stage block counts $(n_1,n_2,n_3,n_4)$ \\
\hline
ResNet--18  & $8$  & $(2,2,2,2)$ \\
ResNet--34  & $16$ & $(4,4,4,4)$ \\
ResNet--50  & $16$ & $(4,4,4,4)$ \\
ResNet--101 & $33$ & $(8,8,8,9)$ \\
ResNet--152 & $50$ & $(12,12,13,13)$ \\
\hline
\end{tabular}
\end{table}

The corresponding epoch allocation is fixed to
\[
(5,\;5,\;30,\;280),
\]
applied to $(\mathcal{S}_1,\mathcal{S}_2,\mathcal{S}_3,\mathcal{S}_4)$ respectively.
At stage $k$, the active backbone consists of the stem and the prefix
$\mathcal{S}_1 \cup \cdots \cup \mathcal{S}_k$.

\vspace{0.4em}
\noindent\textbf{Scenario 2: Two-stage progressive curriculum ($K=2$).}
For the coarse two-stage configuration, the block sequence is divided exactly into
\[
(n_1,n_2) = \text{BalancedPartition}(N,2).
\]
The exact decompositions are shown in Table~\ref{tab:partition-two}.

\begin{table}
\caption{Exact two-stage partitioning used in Scenario~2.}
\label{tab:partition-two}
\begin{tabular}{|l|l|l|}
\hline
Model & $N$ (blocks) & Stage block counts $(n_1,n_2)$ \\
\hline
ResNet--18  & $8$  & $(4,4)$ \\
ResNet--34  & $16$ & $(8,8)$ \\
ResNet--50  & $16$ & $(8,8)$ \\
ResNet--101 & $33$ & $(16,17)$ \\
ResNet--152 & $50$ & $(25,25)$ \\
\hline
\end{tabular}
\end{table}

The training schedule allocates
\[
(50,\;350)
\]
epochs to stages $(\mathcal{S}_1,\mathcal{S}_2)$ respectively.
As in Scenario~1, the backbone at stage~1 contains the stem plus the blocks in
$\mathcal{S}_1$, while stage~2 activates the full backbone.

\vspace{0.4em}
\noindent\textbf{Unified progressive head.}
At every stage, the classifier head remains fixed and consists of
global average pooling, a $1{\times}1$ convolution projecting the output to a
constant-width embedding (256 channels), a ReLU, and a fully connected layer
mapping to the task classes.
Thus, growth occurs solely in the backbone depth, enabling weight reuse and
stable optimization.



\subsection{Progressive Encoder Decomposition for ViT-B/16 and ViT-L/16}

Vision Transformers (ViT-B/16 and ViT-L/16) consist of a sequence of $L$ Transformer
encoder layers:
\[
\mathcal{E} = (e_1, e_2, \ldots, e_L),
\]
with $L=12$ for ViT-B/16 and $L=24$ for ViT-L/16.  
In the proposed framework, each encoder layer $e_i$ is treated as an \emph{atomic computational block}, analogous to a residual block in ResNet.  Progressive training is obtained by grouping these encoder layers into $K$ ordered stages and enabling a strictly increasing subset of the encoder at each stage.

\vspace{0.4em}
\noindent\textbf{Stage construction.}  
Given $L$ encoder layers and a desired number of stages $K$, we divide the encoder into $K$ contiguous groups
\[
\mathcal{T}_1,\;\mathcal{T}_2,\;\ldots,\;\mathcal{T}_K,
\qquad
\mathcal{T}_k = \{e_i, i\in J_k\},
\]
where $(J_1,\ldots,J_K)$ is a balanced partition of $\{1,\ldots,L\}$.  
We denote the stage sizes by
\[
(|J_1|, |J_2|, \ldots, |J_K|),
\]
and these determine exactly how many encoder layers are active during each stage.

During stage $k$, the model consists of the embedding stem, the positional encoding, the class token, all encoder layers in
\[
\mathcal{T}_1 \cup \cdots \cup \mathcal{T}_k,
\]
and the final classification head.  
Thus, deeper encoder blocks are only introduced after earlier ones have been optimized.

\begin{figure}[t]
\centering
\includegraphics[width=\columnwidth]{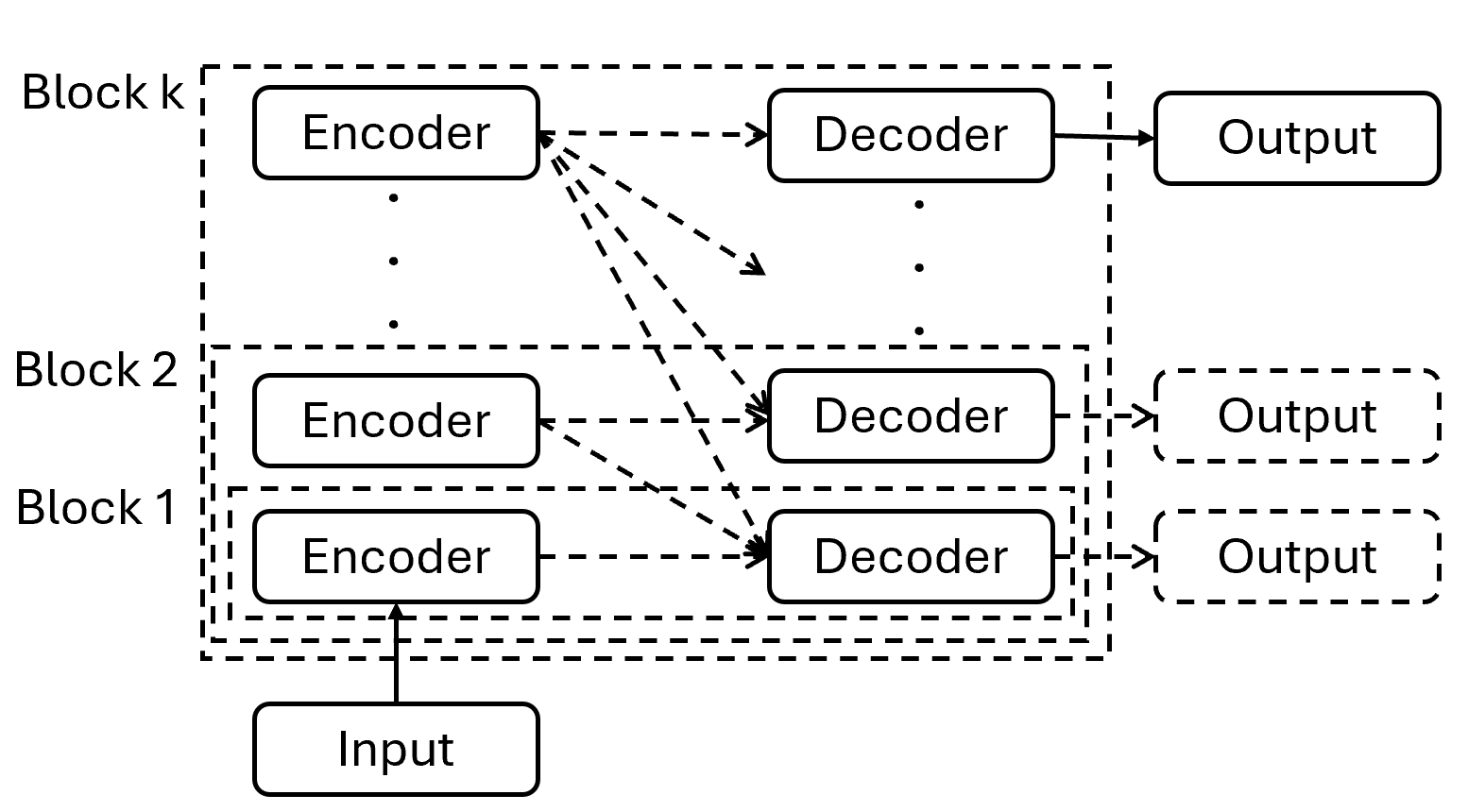}
\caption{Illustration of progressive growing for Transformer encoders: at stage $k$, only the first $k$ blocks are enabled and trained, while deeper blocks remain inactive until later stages.}
\label{fig:progressive_vit_blocks}
\end{figure}

Figure~\ref{fig:progressive_vit_blocks} illustrates this progressive curriculum, where the effective depth grows stage-by-stage while intermediate outputs can optionally be supervised.

\vspace{0.6em}
\noindent\textbf{Exact decomposition used in our experiments.}  
Two progressive configurations were evaluated:

---





\vspace{0.6em}
\noindent\textbf{Scenario 1: $K=2$ progressive stages.}
Epoch allocation: $(50,350)$.
Balanced partitioning yields:

\begin{table}
\caption{Two-stage decomposition for ViT-B/16 and ViT-L/16.}
\label{tab:vit2_exact}
\begin{tabular}{|l|l|l|}
\hline
Model & $L$ & Stage sizes $(|J_1|, |J_2|)$ \\
\hline
ViT-B/16 & $12$ & $(6,6)$ \\
ViT-L/16 & $24$ & $(12,12)$ \\
\hline
\end{tabular}
\end{table}

Stage~1 trains only the earliest half of the encoder, while Stage~2 trains the full set of layers.

\vspace{0.6em}
\noindent\textbf{Unified progressive training principle.}  
This decomposition aligns the ViT architecture with the same progressive framework used for CNN backbones: the network is expressed as a strictly ordered list of atomic blocks, deeper blocks are introduced only after earlier ones converge and training time is distributed unevenly across stages to emphasize the most expressive parts of the network.  This unified formulation makes the progressive curriculum directly comparable between ResNet and ViT architectures.



\section{Results}

\subsection{Progressive Training Performance Analysis}

To further evaluate the effectiveness of progressive training, we analyze its impact across both CNN and Vision Transformer (ViT) backbones, with particular emphasis on the SOS core model, which serves as a representative and structurally sensitive system. The results are summarized in Table~\ref{tab:progressive_full_table}, comparing entire-model training against progressive growing under identical experimental configurations.

For CNN architectures, progressive learning demonstrates a nuanced but meaningful trade-off between accuracy and computational efficiency. In the ResNet-18 configuration under the $(5,5,30,280)$ schedule, the progressive model achieves an accuracy of 0.8451 compared to 0.8603 for the entire model, while requiring only 89.32\% of the total computation. This indicates that progressive training preserves nearly equivalent performance while reducing training cost. Conversely, for ResNet-101, the progressive strategy surpasses the entire-model accuracy (0.8600 vs. 0.8465), highlighting that deeper networks benefit more significantly from staged capacity expansion due to improved gradient stability and hierarchical feature refinement.

A similar pattern emerges for Vision Transformers. The ViT-B/16 baseline shows poor performance under entire-model training (0.4412 accuracy), likely due to optimization instability and insufficient feature hierarchy formation when trained monolithically. However, introducing progressive training results in a marked improvement to 0.6044 accuracy—an absolute gain exceeding 16 percentage points—while consuming only 93.82\% of the total computation. This demonstrates that progressive depth activation is particularly advantageous for transformer models operating on limited or complex morphological datasets such as SOS imagery.

Furthermore, the ViT-L/16 model shows robust entire-model performance (0.6902), suggesting higher baseline capacity.

We also evaluated ViT-B/32 and ViT-L/32 architectures under 50\% patch overlap settings. The entire ViT-B/32 model underperforms with an accuracy of 0.3624, reinforcing the difficulty of direct training on lower-resolution features. However, its progressive counterpart reaches 0.6542 accuracy—outperforming the entire version by a substantial margin. Similarly, ViT-L/32 improves from 0.4924 (entire) to 0.6362 (progressive), confirming that progressive strategies provide optimization stability and performance benefits even for deeper transformer configurations.

Across all evaluated configurations, progressive training consistently delivers either comparable or superior performance relative to standard training, while simultaneously achieving measurable computational savings. More importantly, the gains observed on the SOS system confirm that progressive training enables more structured representation of fine-grained anatomical maturity progression, particularly for stages characterized by subtle morphological transitions.

These results validate progressive growing as both a performance-enhancing and resource-efficient strategy, especially within the context of biologically sensitive classification tasks. The ability to control network complexity dynamically allows for improved convergence behavior, reduced risk of early overfitting, and stronger generalization, reinforcing its suitability for clinical decision-support models.

To verify that the benefits of progressive growing are not limited to SOS data, we additionally report ViT-B/16 results on the public CIFAR-10 benchmark in Table~\ref{tab:progressive_full_table}. Progressive training improves accuracy from 0.7476 to 0.7781 (+3.05\%), with consistent gains in averaged precision/recall/F1, confirming that progressive depth activation improves transformer optimization beyond the clinical SOS setting.

\begin{table}[t]
\caption{Comparison of entire-model and progressive-growing training across CNN and ViT backbones, including 32-patch ViT models with 50\% patch overlap. Standard ResNet-18 training achieves an accuracy of 0.8603 after 320 epocs. Progressive training achieves an accuracy of 0.8617.}
\label{tab:progressive_full_table}
\centering
\scriptsize
\setlength{\tabcolsep}{2.8pt}
\renewcommand{\arraystretch}{1.15}

\newcolumntype{L}[1]{>{\raggedright\arraybackslash}p{#1}}
\newcolumntype{Y}{>{\raggedright\arraybackslash}X}

\begin{tabularx}{\columnwidth}{@{} L{1.55cm} L{2.05cm} L{1.25cm} c Y c @{}}
\toprule
Experiment & Model & Mode & Accuracy & Avg Metrics (Prec / Rec / F1) & Overall computation \\
\midrule
320 & ResNet-18  & Entire      & 0.8603 & 0.8576 / 0.8603 / 0.8581 & --- \\
5, 5, 30, 280 & ResNet-18  & Progressive & 0.8451 & 0.8451 / 0.8451 / 0.8440 & 89.3\% \\
10, 290       & ResNet-18  &  Progressive & \textbf{0.8617} & 0.8599 / 0.8617 / 0.8595 & 96.8\% \\
320 & ResNet-101 & Entire      & 0.8465 & 0.8433 / 0.8465 / 0.8441 & --- \\
5, 5, 30, 280 & ResNet-101 & Progressive & {\bf 0.8600} & 0.8577 / 0.8589 / 0.8565 & 93.2\% \\

\hline
400       & ViT-B/16   & Entire      & 0.4412 & 0.5165 / 0.4412 / 0.4325 & --- \\
50, 350       & ViT-B/16   & Progressive & 0.6044 & 0.6039 / 0.6044 / 0.6012 & 93.8\%  \\
400       & ViT-L/16   & Entire      & 0.5878 & 0.6011 / 0.5878 / 0.5933 & --- \\
50, 350       & ViT-L/16   & Progressive & \textbf{0.6902} & 0.7032 / 0.6902 / 0.6933 & 93.8\%  \\
400       & ViT-B/32 (50\% overlap) & Entire      & 0.6086 & 0.6504 / 0.6086 / 0.6219 & --- \\
50, 350       & ViT-B/32 (50\% overlap) & Progressive & 0.6542 & 0.6499 / 0.6542 / 0.6495 & 93.9\% \\
400       & ViT-L/32 (50\% overlap) & Entire      & 0.5602 & 0.5745 / 0.5602 / 0.5650 & --- \\
50, 350       & ViT-L/32 (50\% overlap) & Progressive & 0.6362 & 0.6467 / 0.6362 / 0.6388 & 93.82\% \\

\hline
\multicolumn{6}{@{}l@{}}{\textit{Public benchmark (CIFAR-10)}}\\
25    & ViT-B/16 & Entire      & 0.7476 & 0.7511 / 0.7476 / 0.7473 &  ---\\
3, 22 & ViT-B/16 & Progressive & \textbf{0.7781} & 0.7800 / 0.7781 / 0.7772 & 94.1\% \\
\bottomrule
\end{tabularx}
\end{table}

\section{Discussion}

Automated assessment of spheno-occipital synchondrosis (SOS) maturation remains a challenging problem due to the subtle and continuous nature of synchondral fusion, the limited availability of annotated CBCT datasets, and pronounced class imbalance across maturation stages. Clinically, SOS fusion progresses gradually rather than discretely, and adjacent stages often exhibit minimal morphological differences. These characteristics make SOS staging particularly sensitive to optimization instability, overfitting, and poor generalization when conventional deep learning models are trained in a single end-to-end manner.

Recent deep learning studies on SOS assessment have demonstrated that convolutional neural networks and transformer-based models can achieve promising accuracy when applied to CBCT-derived sagittal slices. However, prior work consistently reports reduced performance for transitional stages, sensitivity to dataset size, and large variance across training runs. Standard training paradigms that expose the full model capacity from the outset often struggle to reliably learn fine-grained inter-stage distinctions, especially when deeper architectures or high-capacity transformers are employed. These observations suggest that architectural scale alone is insufficient to address the intrinsic difficulty of SOS maturation modeling.

The central contribution of this work is the introduction of a progressive training strategy tailored to SOS maturation assessment. Rather than modifying network architectures or introducing additional supervision, the proposed method restructures the optimization process itself. By incrementally increasing model depth during training, the network is encouraged to first learn coarse, low-level morphological representations of the synchondrosis before gradually incorporating higher-level abstractions required to discriminate adjacent fusion stages. This staged learning process aligns naturally with the biological continuum of SOS fusion and acts as an implicit curriculum without altering class labels or loss functions.

The experimental results across multiple backbones clearly demonstrate the advantages of this approach. For convolutional architectures, progressive training consistently achieves performance comparable to or exceeding that of entire-model training while reducing overall computational cost. Notably, deeper CNNs such as ResNet-101 benefit more substantially from progressive growing, suggesting that staged capacity expansion mitigates gradient instability and overfitting that can arise in deep end-to-end optimization under limited data. These findings indicate that progressive training is particularly effective when model depth increases beyond what the dataset can reliably support in a single training phase.

The benefits of progressive training are even more pronounced for Vision Transformer architectures. Entire-model training of ViT variants exhibits severe optimization instability, leading to poor accuracy and unreliable convergence, especially for smaller or lower-resolution patch configurations. Introducing progressive encoder activation results in substantial performance gains across all transformer settings, with absolute accuracy improvements exceeding 15 percentage points in several cases. These results highlight that transformers, which lack the strong inductive biases of CNNs, are especially sensitive to training dynamics in small medical imaging datasets, and that progressive depth activation provides a practical mechanism for stabilizing representation learning.

Importantly, these gains are achieved without altering preprocessing, loss design, or architectural components, underscoring that training strategy alone can have a decisive impact on SOS classification performance. Progressive training enables more structured representation learning, improves convergence behavior, and enhances discrimination of morphologically adjacent SOS stages—precisely where clinical interpretation is most difficult.

From a clinical perspective, improved reliability in SOS staging directly translates to more consistent assessment of cranial base maturation and growth status. By reducing sensitivity to initialization and overfitting, the proposed framework supports reproducible deployment in clinical decision-support settings, particularly in cases where SOS serves as the primary available maturity indicator due to limited CBCT field of view.

Overall, this study demonstrates that progressive model training is a powerful and generalizable strategy for SOS maturation assessment. Beyond improving accuracy, it offers a principled approach to aligning deep learning optimization with the biological progression of skeletal fusion, making it especially well-suited for anatomically continuous and data-limited classification tasks.

\section{Conclusion}

This study presents a progressive training strategy for automated spheno-occipital synchondrosis (SOS) maturation assessment from CBCT images. By gradually increasing model depth during training, the proposed approach improves optimization stability and stage discrimination under limited and imbalanced data conditions. Experiments across CNN and Vision Transformer backbones show that progressive training consistently matches or outperforms conventional end-to-end training while reducing computational cost. The gains are most pronounced for deeper networks and transformer models, which are particularly sensitive to direct full-capacity optimization. Importantly, these improvements are achieved without modifying network architectures or loss functions, emphasizing the role of training dynamics in SOS classification. The results establish progressive training as a practical and effective strategy for biologically continuous maturity assessment tasks.

\clearpage

%
%
%

\begin{thebibliography}{99}

\bibitem{Mito2002}
Mito, T., Sato, K., Mitani, H.:
Cervical vertebral bone age in girls.
American Journal of Orthodontics and Dentofacial Orthopedics \textbf{122}(4), 380--385 (2002).
\doi{10.1067/mod.2002.126896}

\bibitem{calfee2010skeletal}
Calfee, R.P., Sutter, M., Steffen, J.A., Goldfarb, C.A.:
Skeletal and chronological ages in American adolescents: current findings in skeletal maturation.
Journal of Children's Orthopaedics \textbf{4}(5), 467--470 (2010)

\bibitem{karlberg2002secular}
Karlberg, J.:
Secular trends in pubertal development.
Hormone Research \textbf{57}(Suppl. 2), 19--30 (2002)

\bibitem{de2014hand}
De~Sanctis, V., Di~Maio, S., Soliman, A.T., Raiola, G., Elalaily, R., Millimaggi, G.:
Hand X-ray in pediatric endocrinology: Skeletal age assessment and beyond.
Indian Journal of Endocrinology and Metabolism \textbf{18}(Suppl 1), S63--S71 (2014)

\bibitem{Scarfe2017}
Scarfe, W.C., Azevedo, B., Toghyani, S., Farman, A.G.:
Cone beam computed tomographic imaging in orthodontics.
Australian Dental Journal \textbf{62}, 33--50 (2017)

\bibitem{OlchAlaei2021}
Olch, A.J., Alaei, P.:
How low can you go? A CBCT dose reduction study.
Journal of Applied Clinical Medical Physics \textbf{22}(2), 85--89 (2021)

\bibitem{alhazmi2021correlation}
Alhazmi, A., Aldossary, M., Palomo, J.M., Hans, M., Latimer, B., Simpson, S.:
Correlation of spheno-occipital synchondrosis fusion stages with a hand-wrist skeletal maturity index:
a cone beam computed tomography study.
The Angle Orthodontist \textbf{91}(4), 538--543 (2021)

\bibitem{booth2024correlations}
Booth, E., Viana, G., Shirazi, S., Miller, S., Sellke, T., Elnagar, M., Viana, M., Atsawasuwan, P.:
Correlations of spheno-occipital synchondrosis, cervical vertebrae, midpalatal suture,
and third molar maturation stages.
Angle Orthodontist \textbf{94}(6), 641--647 (2024).
\doi{10.2319/041224-295.1}

\bibitem{fernandez2016spheno}
Fern{\'a}ndez-P{\'e}rez, M.J., Alarc{\'o}n, J.A., McNamara~Jr., J.A., Velasco-Torres, M.,
Benavides, E., Galindo-Moreno, P., Catena, A.:
Spheno-occipital synchondrosis fusion correlates with cervical vertebrae maturation.
PLoS ONE \textbf{11}(8), e0161104 (2016)

\bibitem{powell1963closure}
Powell, T.V., Brodie, A.G.:
Closure of the spheno-occipital synchondrosis.
The Anatomical Record \textbf{147}(1), 15--23 (1963)

\bibitem{lottering2015ontogeny}
Lottering, N., MacGregor, D.M., Alston, C.L., Gregory, L.S.:
Ontogeny of the spheno-occipital synchondrosis in a modern Queensland, Australian population using computed tomography.
American Journal of Physical Anthropology \textbf{157}(1), 42--57 (2015)

\bibitem{krishan2013evaluation}
Krishan, K., Kanchan, T.:
Evaluation of spheno-occipital synchondrosis: A review of literature and considerations from forensic anthropologic point of view.
Journal of Forensic Dental Sciences \textbf{5}(2), 72--76 (2013)

\bibitem{tashayyodi2023relationship}
Tashayyodi, N., Kajan, Z.D., Ostovarrad, F., Khosravifard, N.:
Relationship of the Fusion Stage of Spheno-Occipital Synchondrosis with Midpalatal and Zygomaticomaxillary Sutures
on Cone-Beam Computed Tomography Scans of Patients Aged Between 7 and 21 Years.
Turkish Journal of Orthodontics \textbf{36}(3), 186 (2023)

\bibitem{alhazmi2017timing}
Alhazmi, A., Vargas, E., Palomo, J.M., Hans, M., Latimer, B., Simpson, S.:
Timing and rate of spheno-occipital synchondrosis closure and its relationship to puberty.
PLoS ONE \textbf{12}(8), e0183305 (2017)

\bibitem{leonardi2010rapid}
Leonardi, R., Cutrera, A., Barbato, E.:
Rapid maxillary expansion affects the spheno-occipital synchondrosis in youngsters:
a study with low-dose computed tomography.
The Angle Orthodontist \textbf{80}(1), 106--110 (2010)

\bibitem{leonardi2021three}
Leonardi, R., Ronsivalle, V., Lagravere, M.O., Barbato, E., Isola, G., Lo~Giudice, A.:
Three-dimensional assessment of the spheno-occipital synchondrosis and clivus after tooth-borne and bone-borne rapid maxillary expansion:
a retrospective CBCT study using voxel-based superimposition.
The Angle Orthodontist \textbf{91}(6), 822--829 (2021)

\bibitem{goldstein2014earlier}
Goldstein, J.A., Paliga, J.T., Wink, J.D., Bartlett, S.P., Nah, H.-D., Taylor, J.A.:
Earlier evidence of spheno-occipital synchondrosis fusion correlates with severity of midface hypoplasia
in patients with syndromic craniosynostosis.
Plastic and Reconstructive Surgery \textbf{134}(3), 504--510 (2014)

\bibitem{tahiri2014spheno}
Tahiri, Y., Paliga, J.T., Vossough, A., Bartlett, S.P., Taylor, J.A.:
The spheno-occipital synchondrosis fuses prematurely in patients with Crouzon syndrome and midface hypoplasia compared with age-
and gender-matched controls.
Journal of Oral and Maxillofacial Surgery \textbf{72}(6), 1173--1179 (2014)

\bibitem{Kapila2015}
Kapila, S.D., Nervina, J.M.:
CBCT in orthodontics: Assessment of treatment outcomes and indications for its use.
Dentomaxillofacial Radiology \textbf{44} (2015).
Preprint at \url{https://doi.org/10.1259/dmfr.20140282}

\bibitem{lee2017deep}
Lee, J.-G., Jun, S., Cho, Y.-W., Lee, H., Kim, G.B., Seo, J.B., Kim, N.:
Deep learning in medical imaging: general overview.
Korean Journal of Radiology \textbf{18}(4), 570--584 (2017)

\bibitem{MohammadRahimi2021}
Mohammad-Rahimi, H., Nadimi, M., Rohban, M.H., Shamsoddin, E., Lee, V.Y., Motamedian, S.R.:
Machine learning and orthodontics, current trends and the future opportunities: A scoping review.
American Journal of Orthodontics and Dentofacial Orthopedics \textbf{160}(2), 170--192.e4 (2021).
\doi{10.1016/j.ajodo.2021.02.013}

\bibitem{zhang2020automatic}
Zhang, J., Li, C., Song, Q., Gao, L., Lai, Y.-K.:
Automatic 3D tooth segmentation using convolutional neural networks in harmonic parameter space.
Graphical Models \textbf{109}, 101071 (2020)

\bibitem{thurzo2021use}
Thurzo, A., Kosn{\'a}{\v{c}}ov{\'a}, H.S., Kurilov{\'a}, V., Kosmel', S., Be{\v{n}}u{\v{s}}, R.,
Moravansk{\`y}, N., Kov{\'a}{\v{c}}, P., Kuracinov{\'a}, K.M., Palkovi{\v{c}}, M., Varga, I.:
Use of advanced artificial intelligence in forensic medicine, forensic anthropology and clinical anatomy.
In: Healthcare, vol. 9(11), p.~1545 (2021)

\bibitem{kumar2024cbct}
Kumar, N., Malik, B., Dubey, A., Kaur, H., Mujoo, S., Jugade, S.C., Gulia, S.K.:
Long-Term Trends in the Utilization of Cone Beam Computed Tomography in Oral and Maxillofacial Radiology.
Journal of Pharmacy and Bioallied Sciences \textbf{16}(Suppl 3), S2585--S2587 (2024).
\doi{10.4103/jpbs.jpbs_295_24}

\bibitem{bassed2010analysis}
Bassed, R.B., Briggs, C., Drummer, O.H.:
Analysis of time of closure of the spheno-occipital synchondrosis using computed tomography.
Forensic Science International \textbf{200}(1--3), 161--164 (2010).

\bibitem{AticiCVMDirectionalFilters}
Atici, S.F., Ansari, R., Allareddy, V., Suhaym, O., Cetin, A.E., Elnagar, M.H.:
Fully automated determination of the cervical vertebrae maturation stages using deep learning with directional filters.
PLOS ONE \textbf{17}(7), e0269198 (2022)


\end{thebibliography}
%





\end{document}